\documentclass{article} 
\usepackage{iclr2025_conference,times}

\usepackage{amsmath,amsfonts,bm}

\def\eqref#1{equation~\ref{#1}}

\def\1{\bm{1}}

\DeclareMathAlphabet{\mathsfit}{\encodingdefault}{\sfdefault}{m}{sl}
\SetMathAlphabet{\mathsfit}{bold}{\encodingdefault}{\sfdefault}{bx}{n}

\usepackage{hyperref}
\usepackage{url}
\usepackage{tabularray}

\usepackage{graphicx} 

\title{BarkXAI: A Lightweight Post-Hoc Explainable Method for Tree Species Classification with Quantifiable Concepts}

\author{Yunmei Huang\thanks{These authors contributed equally to this work.} \\
Department of Forestry and Natural Resources\\
Purdue University\\
West Lafayette, IN, USA \\
\texttt{huan1643@purdue.edu} \\
\And
Songlin Hou$^*$ \\
Department of Computer Science\\
Worcester Polytechnic Institute\\
Worcester, MA, USA\\
\texttt{shou@wpi.edu} \\
 \AND
Zachary Nelson Horve \\
Department of Forestry and Natural Resources \\
Purdue University \\
West Lafayette, IN, USA \\
\texttt{zhorve@purdue.edu} \\
\And
Songlin Fei \\
Department of Forestry and Natural Resources \\
Purdue University \\
West Lafayette, IN, USA \\
\texttt{sfei@purdue.edu} 
}

\iclrfinalcopy 
\begin{document}

\maketitle

\begin{abstract}
The precise identification of tree species is fundamental to forestry, conservation, and environmental monitoring. Though many studies have demonstrated that high accuracy can be achieved using bark-based species classification, these models often function as "black boxes", limiting interpretability, trust, and adoption in critical forestry applications. Attribution-based Explainable AI (XAI) methods have been used to address this issue in related works. However, XAI applications are often dependent on local features (such as a head shape or paw in animal applications) and cannot describe global visual features (such as ruggedness or smoothness) that are present in texture-dominant images such as tree bark. Concept-based XAI methods, on the other hand, offer explanations based on global visual features with concepts, but they tend to require large overhead in building external concept image datasets and the concepts can be vague and subjective without good means of precise quantification. To address these challenges, we propose a lightweight post-hoc method to interpret visual models for tree species classification using operators and quantifiable concepts. Our approach eliminates computational overhead, enables the quantification of complex concepts, and evaluates both concept importance and the model’s reasoning process. To the best of our knowledge, our work is the first study to explain bark vision models in terms of global visual features with concepts. Using a human-annotated dataset as ground truth, our experiments demonstrate that our method significantly outperforms TCAV and Llama3.2 in concept importance ranking based on Kendall’s Tau, highlighting its superior alignment with human perceptions.
\textit{Source code will be released upon acceptance.}


 
\end{abstract}

\section{ Introduction }
%

Forests play a crucial role in climate solutions by sequestering carbon, regulating ecosystems, and supporting biodiversity. Tree species identification, as an essential means of forest management, helps understanding forest composition, and ecosystems. With the rapid advancement in AI, tree species identification automation with deep neural networks is increasingly gaining attention. Many studies \citep{carpentier2018tree_cnn_bark_id,bertrand2018bark_leaf_fusion,robert2020tree_deepBarkdataset,Wu2021Bark_wufanyou,yamabe2022vision_barkid} have demonstrated that high accuracy in tree species identification can be achieved using bark images, which are available year-around compared with leaf, flower, and fruits. However, most of the existing studies fall short of explaining the reasoning process behind their models. The inherent complexity of these models often renders them black boxes, making it difficult to understand. This lack of transparency can hinder trust and transferability in the model's output, especially in critical applications like forestry, where accurate and reliable species identification is important~\citep{onishi2021explainable_uav,cheng2022improve_model_tree_explation,hohl2024recent_trend_remotesensing_explainable_ai}.


Considering the intra- and inter-species complexity of tree bark images, understanding deep neural networks for tree species classification is important to unveil the black box and provide insights for future applications. 
To address these challenges, Explainable AI (XAI) has emerged to enhance transparency and trust in deep neural networks by shedding light on their reasoning processes~\citep{mostafa2023explainable_phenotypeing_reivew}.
Existing XAI methods can be broadly categorized into attribution-based XAI (which focus on attributing importance to input features that drive model decisions) and concept-based XAI (which interprets models using high-level human-understandable concepts).

Attribution-based XAI methods such as Crown-CAM~\citep{marvasti2023crown__cam}, and Grad-CAM  \citep{onishi2021explainable_uav,ahlswede2022weakly_cam,kim2022identifying_bark_cam,huang2024temperate_gradcam} have been used to visualize which image regions contribute to classification decisions in aerial and ground imagery and improve the explainability of tree species identification. Similarly, Shapeley Additive explanations (SHAP)~\citep{lundberg2017unified}, have been utilized in species richness modeling\citep{brugere2023improved_richness_Explainablity}, urban vegetation mapping \citep{abdollahi2021urban_sheap_XAI} and mulberry leaf disease classification~\citep{nahiduzzaman2023explainable_mulberry}. Local Interpretable Model-Agnostic explanations(LIME) has been applied to provide localized visual interpretations in medicinal plant species identification~\citep{nikam2022explainable_spp_lime} and microscopic wood classification~\citep{zhan2023wood_id_lime}.\\

\begin{figure}
    \centering
    \includegraphics[width=\linewidth]{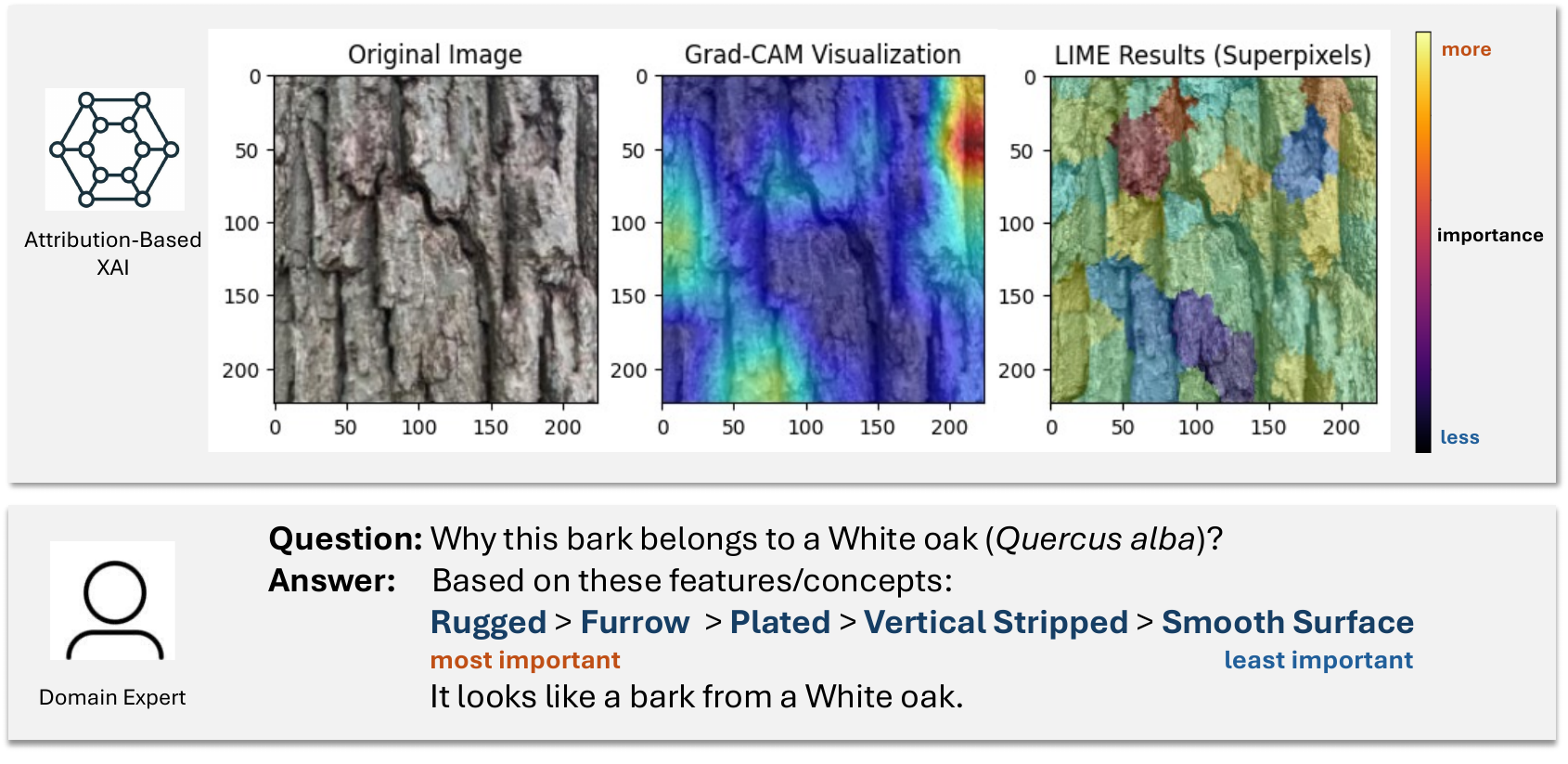}
    \caption{Illustration of Attribution-Based Methods on a Bark Image (Trained and Evaluated with MobileNetV2) and domain expert explanation. While attribution-based methods are widely used in bark vision model explanation and similar fields, they cannot reveal global visual features used by domain experts.}
    \label{fig:attr_xai_vs_human}
\end{figure}

While attribution-based methods are more popular in explaining vision models, they usually cannot provide satisfactory results when interpreting vision classifiers on bark images due to a lack of prominent local features. Attribution-based methods work best on natural images containing distinct local features such as animals (head, paws, tails) and vehicles (wheels, doors) by providing explanations that highlight these local features, which are easily verifiable by users. However, for texture-dominant images that lack prominent local features, such as tree bark, these methods struggle to provide explanations that align with human reasoning. Contrary to the explanation from attribution-based methods, domain experts typically rely on global visual characteristics—such as stripe patterns, roughness, and surface irregularities—to distinguish tree species, which are not revealed by attribution-based methods. The gap in explanation is further illustrated in Figure \ref{fig:attr_xai_vs_human}.

In contrast, concept-based XAI~\citep{kim2018interpretability_TCAV, chen2019looks} methods offer a more intuitive approach to interpreting models using high-level, human-understandable concepts to explain model decisions, which aligns better with our intuition when distinguishing tree species based on bark images\citep{kazhdan2020now_CME_CXAI,sagar2023leaf_xai}. However, extant concept-based models typically require pre-defined concepts and external effort in collecting images for each concept, which quickly becomes labor-extensive when the number of concepts is large\citep{kim2018interpretability_TCAV,hou2024conceptXAI_skin}. One exception is ProtoPNet\citep{chen2019looks}, which eliminates the requirements of collecting external concept images by using image patches from training set as prototypes to provide concept-based explanations. However, the way image patches are used as concepts still limits the ability to describe global visual features, like attribution-based methods. Besides, concepts that exhibit a clear hierarchical relationship (e.g., light red vs. deep red) or require precise quantification (e.g., 30-degree vs. 60-degree inclination) pose significant challenges for existing concept-based methods since accurately defining and representing such concepts using external concept images is vague and subjective.


To provide better explanation on bark visual models that matches the intuition of domain experts while mitigating the limitations discussed above, we propose BarkXAI, which is a lightweight post-hoc concept-based method optimized for bark image classifiers. To the best of our knowledge, few, if any, of the studies have adopted concept-based XAI methods when interpreting vision models on bark images. \textit{Our work is the first to provide a concept-based explanation on global visual features on vision models on bark image or texture-dominant images identification.} Our main contributions are listed as follows.

\begin{itemize}
    \item By using operators instead of concept images to reflect features related to bark images, we eliminate the overhead of collecting external concept images found in traditional concept-based XAI methods while ensuring greater practicality.
    \item We provide a way to construct quantifiable concepts that are hard to define and formalize in existing concept-based XAI approaches. 
    \item Our method can evaluate concept importance as well as the reasoning process with inter-concept relationships.
\end{itemize}

\section{BarkXAI: proposed method}
\label{BarkXAI}
 
\begin{figure}
    \centering
    \includegraphics[width=1\linewidth]{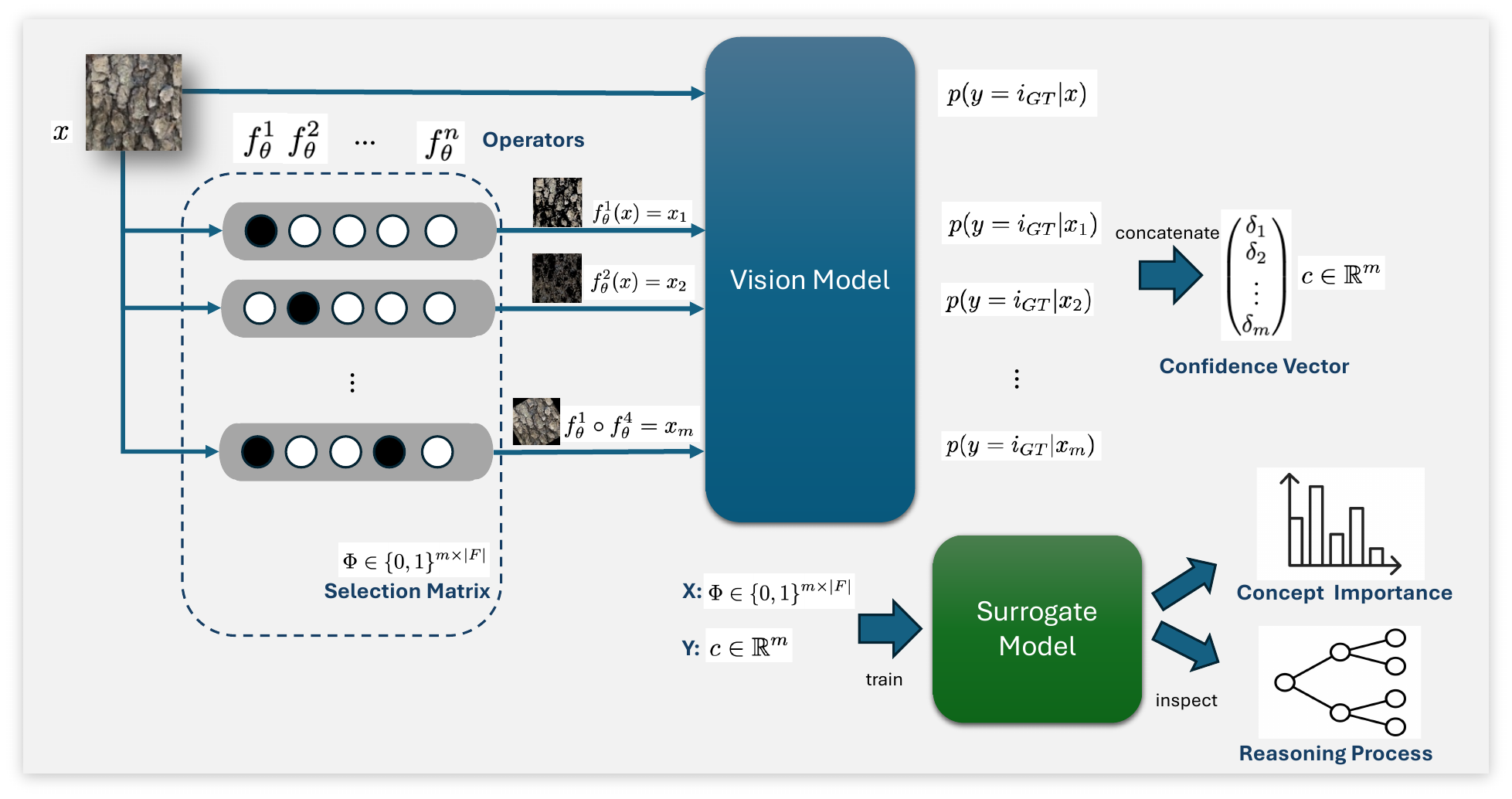}
    \caption{Pipeline of BarkXAI in Explaining Bark Vision Models}
    \label{fig:explaining_process}
\end{figure}

Given the limitations in existing XAI models for explaining vision models, we aim to address how to design an XAI method that:  
1) Can explain any trained black-box texture vision models,  
2) Uses quantifiable concepts meaningful for texture analysis, and  
3) Minimizes additional computational overhead.  
To solve this, we propose BarkXAI, a novel XAI approach optimized for image classifiers trained on texture images (e.g., bark images). Inspired by LIME and TCAV~\citep{kim2018interpretability_TCAV}, our method uses perturbed images to assess the impact of concept-of-interest on classification. Unlike LIME, which relies on superpixel segmentation, we focus on global visual features (e.g., smoothness or tone) that are visually significant for texture analysis. These features are extracted and perturbed, with impacts analyzed using surrogate models like linear regression or decision trees. Similar to TCAV, our method explains model decisions through concepts but uses parameterized operators instead of externally curated datasets, enabling efficient concept evaluation without extensive dataset preparation. Our proposed approach is illustrated in Figure~\ref{fig:explaining_process}.

\subsection{Global Visual Features}
Global visual features in images refer to attributes or descriptors that depict the overall structure or appearance of an entire image rather than concentrating on specific objects or localized regions. These features encompass high-level concepts and they can be categorized into the following types.
\begin{itemize}
    \item  Color: Features of colors can be reflected in several ways, such as color histogram and color correlograms. They reflect the distribution or correlation of pixel colors across the image. For bark images, color can be an important indicator for species classification.
    \item Texture: Patterns or surface properties, such as smoothness, coarseness, or regularity, are utilized to characterize textures. In conjunction with variations in illumination conditions, smoothness, and coarseness can provide insights into the evenness or roughness of the surface. In the case of bark images, the presence of ridges and cracks serves as important indicators of tree species. These features also contribute to estimating the age and health condition of the trees. 
    \item Shape: The geometric structure of the texture surface can be described by properties such as lines, edges, and circles, among others. The dominant orientation or directionality, along with the presence of these features, provides valuable insights into the texture type.
    \item Groove and surfaces: Grooves in tree bark typically exhibit significant visual activity that can be meaningful to species classification. More complex surfaces are characterized by a greater number of edges, variations, and detail while simpler surfaces tend to display fewer visual elements and greater uniformity. 
\end{itemize}

\begin{table}
\centering
\caption{Operators (Concept Key-Value Pairs)}
\label{Tab:opts}
\begin{tblr}{
  cell{4}{1} = {r=2}{},
  cell{6}{1} = {r=2}{},
  hline{1-2,8} = {-}{},
}
\textbf{Concept}  & \textbf{Key} & \textbf{Value}      & \textbf{Notes}                                                      \\
Color             & Tune         & +5/+10              & Overall color tune is increased by 30 and ~50      \\
Texture           & Smooth       & +150/~+300       & Smooth image (+150/ +300)                                           \\
Shape ~           & Flip         & Horizontal/Vertical & Flip image horizontally/vertical                                    \\
                  & Rotate       & +$\pm30$/$\pm90$      & Rotate image by$\pm30$, $\pm90$ (clockwise/anticlockwise) \\
Groove and Surface & Groove         & Remove              & Remove the groove parts of the image.                                \\
                  & Surface      & Remove              & Remove the surface parts of the image.                              
\end{tblr}
\end{table}



\subsection{Commutative Operators}

To extract global visual features from texture images, we introduce a collection of operators $F$, which serve as feature extractors. Each operator $f \in F$ is generally designed as an unary function, which takes one input $x$ and outputs $x'$, which is a perturbed version of $x$. For some operators $f\in F$, there might be a parameterized version that can take a vector of parameters such as $f_{\theta}$.

Each of the operators ($f \in F$) is designed to involve exactly one concept key-value pair. The concept refers to the type of global visual feature being evaluated, while the key-value pairs under each concept represent specific values associated with that concept. We summarize the concept key-value pairs in Table~\ref{Tab:opts}. Note that the concept key-value pairs used in our method do not constitute a comprehensive list for describing global visual features, as it is impractical to enumerate and evaluate all possible concepts for each visual feature. Moreover, the classification of key-value pairs can be ambiguous, given that the aspects of concepts used to describe texture images often overlap.  

\subsubsection{Color (Tune) Operator}
The color operator is designed to manipulate the hue component of an image,  altering its color characteristics. This operator takes an image input, which is then transformed into the HSV (Hue, Saturation, Value) color space. Applying the HSV conversion allows us to separate the chromatic content (hue) from the image, allowing for pixel-intensity-agnostic manipulation.

We adjust the first channel (which is the hue channel) by adding a predefined parameter representing the changes in hue. The adjustment is performed modulo 180 (range of hue value) to ensure that the updated hue values remain within valid bounds. The adjusted image is converted back to the RGB color space to keep consistency with the original input image. We use the standard deviation ($\sigma$) of hue values from all bark images as a standard unit to quantify the variation of hue values. In experiments, we increase the tune value by 5 (close to $1 \times \sigma$) and 10 (close to $2 \times \sigma$), to simulate the color changes in texture images.

\subsubsection{Smooth Operator}



The smooth operator refines the texture image surface. While various smoothing algorithms, such as Gaussian and median blur, often blur edges and obscure geometric structures, a bilateral filter is used to preserve edge integrity. This filter achieves edge-preserving smoothing by considering both spatial proximity and pixel intensity differences, with adjustable sensitivity to control smoothness levels. Two sensitivity values ($+150, +300$) are tested, affecting the degree of smoothing and potential texture detail loss. The results are visually inspected to ensure integrity.

\subsubsection{Groove/Surface Removal Operator}
A  groove is a small valley structure found on the surface of stems, branches, and other plant organs of woody plants; for bark images, they are roughly divided into grooves and surface areas for ease of evaluation. Compared to the surface areas, which are generally smoother, grooved areas are spongier or more porous, with a slightly rougher or raised texture relative to the surrounding regions. The differences in shape often result in color variation due to uneven illumination, and the edges with the highest color contrast are identified as the contours of the grooved area.

While accurate segmentation of the grooved area should be ideal for perturbing images accurately, it defeats our purpose of building a simple and intuitive operator for image explanation with extra human effort. Additionally, the accurate segmentation of grooves, even if manageable, does not improve the explainability in our proposed solution. Instead, we propose and implement a simple vision-based pipeline to segment the grooved areas based on the color contrastness of the texture images. 
\subsection{Process to explain image }
The pipeline consists of four steps. (1) First, it begins by converting the input image to a grayscale representation to simplify the image data as a way to reduce computational complexity while preserving essential structural information. (2) The resultant image is thresholded to separate the foreground from the background, resulting in a binary image. To determine an optimal threshold value automatically, we employ Otsu's thresholding method which adaptively determines the threshold value. 3) We apply morphological operations to enhance feature detection by removing noises and filling small gaps. With these two operations, only the significant features can be retained. (4) Lastly, we scan the image and identify all the contours, which are the boundaries of connected components in the image, returning a list of contour points.

The contours which correspond to the significant features in the image are used as binary masks to segment the area of the grooves. Conversely, we can segment the surface area using a flipped mask generated by applying logical NOT operation on the groove mask. Figure~\ref{chart_barkXAI} shows the intermediate results of each step. The figure shows the results after processing the groove- and surface-only images using seam carving, which can further demonstrate the effectiveness of our proposed pipeline.

\subsubsection{Shape Operators}


 Flipping and rotating images effectively alter the directionality of dominant texture features while preserving dimensional consistency. However, $\pm90$-degree rotations may cause pixel loss due to clipping in non-square images, and $\pm30$-degree rotations can introduce black pixels. Flipping shifts dominant features to the opposite side, and when combined with rotation, helps assess the impact of feature positioning on vision models.

\subsubsection{Commutativity under Composition Operations}

Commutativity under Composition Operations (\textbf{CUCO}) is a property defined for a set of operators $F$. Let $f: X \to X$  and $g: X \to X$  be two operators/functions defined on a set $X$. The operation of composition of these functions, denoted  $(f \circ g)(x) = f(g(x))$ , is said to be commutative if and only if:

\begin{equation}
    f \circ g = g \circ f
\end{equation}

That is, for all  $x \in X$ ,

\begin{equation}
    f(g(x)) = g(f(x))
\end{equation}

\textbf{CUCO} property is desired for the operators in our settings because it allows us to study the independent influence of each operator on the vision model without accounting for the order of their application. This simplifies the modeling process by eliminating the need to consider inter-correlations between operators based on sequence order. Specifically, for any combination of the mentioned operators, we expect the final result to be unique and independent of the order in which the operators are applied. While it may not be feasible for our designed operators to fully satisfy this property due to unavoidable information loss in certain operations, such as rotation, operators that partially satisfy \textbf{CUCO} remain valuable. Furthermore, testing this property theoretically can be impractical given the dissimilarity in the internal procedures of each operator, so we employ a numerical approach to assess the degree to which our operators achieve the \textbf{CUCO} property. 

In our experiment, we calculate the average MAE value between images (with pixel range $0-255$) generated with same set of operators but applied in different orders. The average MAE across our dataset is less than $30$, which indicates the operators we defined partially satisfy this property.

\subsection{Explaining with Surrogate Models}
\begin{figure}[h]
\begin{center}
\includegraphics[width=0.6\textwidth]{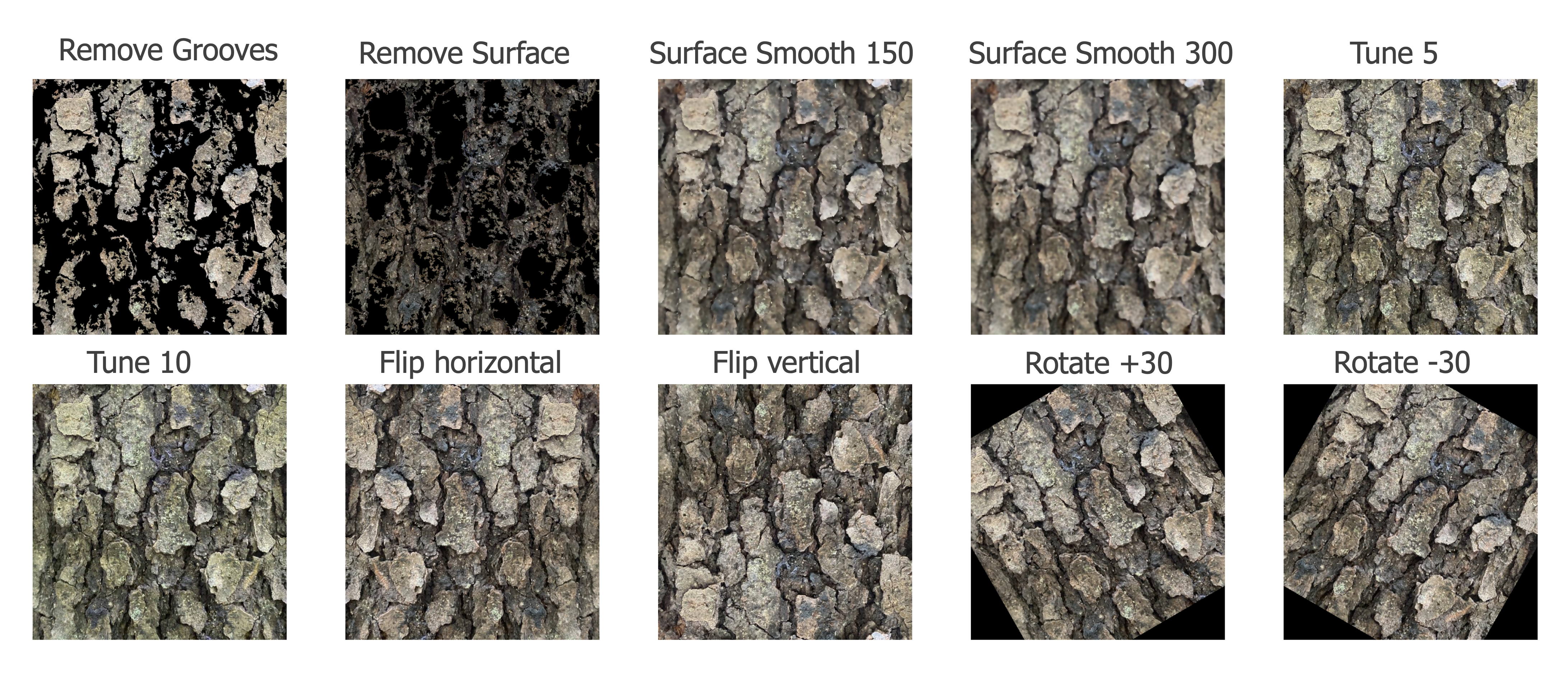}  
  \caption{Ten operations demonstration on Black cherry~(\textit{Prunus serotina}). }
\label{chart_barkXAI_operation}
\end{center}
\end{figure}

We randomly sample a sequence of operators $f_1, f_2, ..., f_N$ from the set of all defined operators $f \in F$ and apply them to perturb the input image $x$. The perturbed image, denoted as $x' = f_1 \circ f_2 \circ \cdots \circ f_N(x)$, is obtained after sequentially applying all the operators. The image $x'$ is then used as input to the vision model, and the confidence value (probability) $p(y = i_{GT} | x')$, corresponding to the class ID $i_{GT}$ in the output probability distribution (typically generated by a softmax activation function), is recorded. Operators included in the sampled sequence are labeled as 1, while those excluded are labeled as 0, resulting in a binary vector $\varphi \in \{0, 1\}^{|F|}$, where $|F|$ denotes the cardinality of the operator set.
The sampling process is repeated for $m > |F|$ iterations to ensure that every operator is selected at least once. To avoid redundancy, the confidence value is computed only for unique sequences, regardless of the order of the operators. The confidence values are concatenated to form a vector $c \in \mathbb{R}^m$, while the binary vectors for all sampled sequences are concatenated to construct a 2D binary matrix $\Phi \in \{0, 1\}^{m \times |F|}$.
The selection matrix $\Phi$ and the confidence vector $c$ are utilized to train a surrogate model. Both linear regression and Classification and Regression Tree (CART) models are employed to analyze the impact of operator combinations on the vision model's confidence values. Linear regression identifies the independent effect of each operator, while CART explores interactions between operators, providing a comprehensive analysis of the causal relationship between operator usage and the vision model's performance. This process is illustrated in Figure \ref{chart_barkXAI_operation}.

With these surrogate models, we can interpret the impact of operators on the confidence value as the influence of the concepts behind the operators to the performance of the models. For example, if there is a drastic confidence value change after a color operator is used, we can interpret this behavior as \textit{for this image $x$ which has class $c$, color plays a significant role when the vision model classifies it to be in class $c$}. Together with manual observation and domain knowledge, we can verify the correctness of the model and analyze/debug the model when the decision process is against our intuition. We will showcase real examples of explaining vision models in the experiment section.

\section{Experiment results: Understand bark images with XAI}
\subsection{Data Collection and BarkXAI Evaluation}

We built a tree species dataset comprising bark images from 21 species (Table~\ref{21species_list}), with 8,184 images in total.
Images were collected by dendrologists in Indiana, with 3 to 5 photos of each tree from different distances and perspectives, ranging from 1 to 4 meters from the tree trunks. Due to the diversity of species present in both natural forests and plantations, the number of images per species varies but each class contains at least 80 images in the dataset. We further integrated an automated pipeline using YOLOv11~\citep{Jocher_Ultralytics_YOLO_2023} and the Segment Anything Model~\citep{kirillov2023segment}, which includes tree detection and bark segmentation. With the clean bark images (examples in Figure~\ref{chart_barkXAI_operation}), we trained MobileNetv2, a light-weighted deep neural network, to classify tree species only using bark images, with an accuracy of 90+\%. \\
To interpret how models predict tree species, we used all operators (Table~\ref{Tab:opts}) and evaluated how each operator impacts the model's decision process. 
In Figure~\ref{chart_barkXAI}, we illustrate the feature importance of various operators applied on the model in all images. For the vision model, MobileNetV2, "remove grooves" and "remove surface" emerge as the most influential operations affecting species prediction. Meanwhile, different rotation angles produced similar effects on both models.  



\subsection{Performance Comparison}



To quantitatively compare our proposed method with others in providing explanations that align with human intuition, we created a test dataset with the reserved bark images and labeled concepts (21 classes with 20 images in each class). For each image, five concepts (rugged, plated, furrow, vertical stripped, and smooth) were ranked by human visual interpretation as ground truth based on how their importance in identifying tree species. Each concept-based method then predicted the ranking of each concept for every image. The predicted orderings were compared with the ground truth using Kendall’s Tau.

Notably, while our method allows for quantifiable concepts by adjusting parameters, we need to compromise by using limited concepts that are easier to define in most concept-based methods such as TCAV. Similar to other variants, TCAV relies solely on collected concepts images and quantifiable concepts are thus hard to define. Some of the concepts, such as color, are also not well-defined concept images, making quantifiable concepts difficult to define. Some concepts, such as color, are not well-defined within TCAV, complicating the collection of concept images. Therefore, for the propose of fair comparison, we only selected concepts that are generally applicable to concept-based methods. the purpose of a fair comparison, we only pick concepts which are easy to be used in general concept-based methods. 

We prepared externally collected concept images for each concept (100 images for each concept) and the average magnitude of the last 3 feature layers is used as the metric to rank each concept given a test image during TCAV evaluation. We also use llama3.2-vision~\citep{dubey2024llama} to provide rank based on each bark image (zero-shot style) depending on its multi-modal ability. There is no direct mapping between our defined operators to test concepts, so the importance of test concepts are inferred with sensitivities from one or more operators. The same test dataset is used in all methods in the experiment and except for Llama3.2-vision, all the XAI methods are evaluated using the pretrained MobileNetv2 model we discussed earlier. We include further discussion in the appendix.

Table \ref{Tab:tau_comparision} shows the average $-1 \leq \tau \leq 1$ value in each species from several methods, where values close to 1 indicates strong agreement and values close to -1 indicate strong disagreement. BXAI (DT), BXAI(LR), and BXAR(RF) are all variants of our proposed method but with different surrogate models (decision tree, linear regression and random forest respectively). We can observe that nearly half of the predictions from TCAV and zero-shot llama3.2 are against human intuition ($\tau < 0$), and almost all BXAI methods provide importance rankings that agree with ground truth results ($\tau > 0$). Compared with llama3.2, TCAV achieves slightly better performance, however, our proposed method outperformed both by large margins. This indicates that our proposed method, compared with TCAV and llama3.2, gives better model explanations that conform to human understanding. We also include examples of explanation in Section \ref{explanation_example} as well as examples of using decision tree as surrogate model to visualize reasoning process in Figure \ref{chart_barkXAI_TREE} in appendix. 






\section{Conclusion}
In this paper, we propose a lightweight post-hoc XAI method that provides human-interpretable explanations for tree bark vision models, enhancing transparency and facilitating expert validation. Compared to other similar works in vision model explanation, our approach utilizes concepts rather than the commonly used attribution-based methods to explain the global visual features in tree barks. While other concept-based XAI methods typically require building external concept image datasets, we employ operators to create quantifiable concepts without additional overhead. Compared to TCAV and Llama3.2, our method offers explanations that align more closely with human perception. Although our approach demonstrates superior explainability for bark images, we acknowledge that its performance can heavily depend on carefully designed operators, which may require expert knowledge for adaptation across different domains. Furthermore, similar to other concept-based XAI methods, while our approach effectively captures global visual features that are difficult to achieve with attribution-based methods, the explanations remain susceptible to subjectivity.




\bibliography{iclr2025_conference}
\bibliographystyle{iclr2025_conference}

\appendix
\section{Appendix}
 
 \begin{figure}[h]
 \begin{center}
 \includegraphics[width=0.9\textwidth]{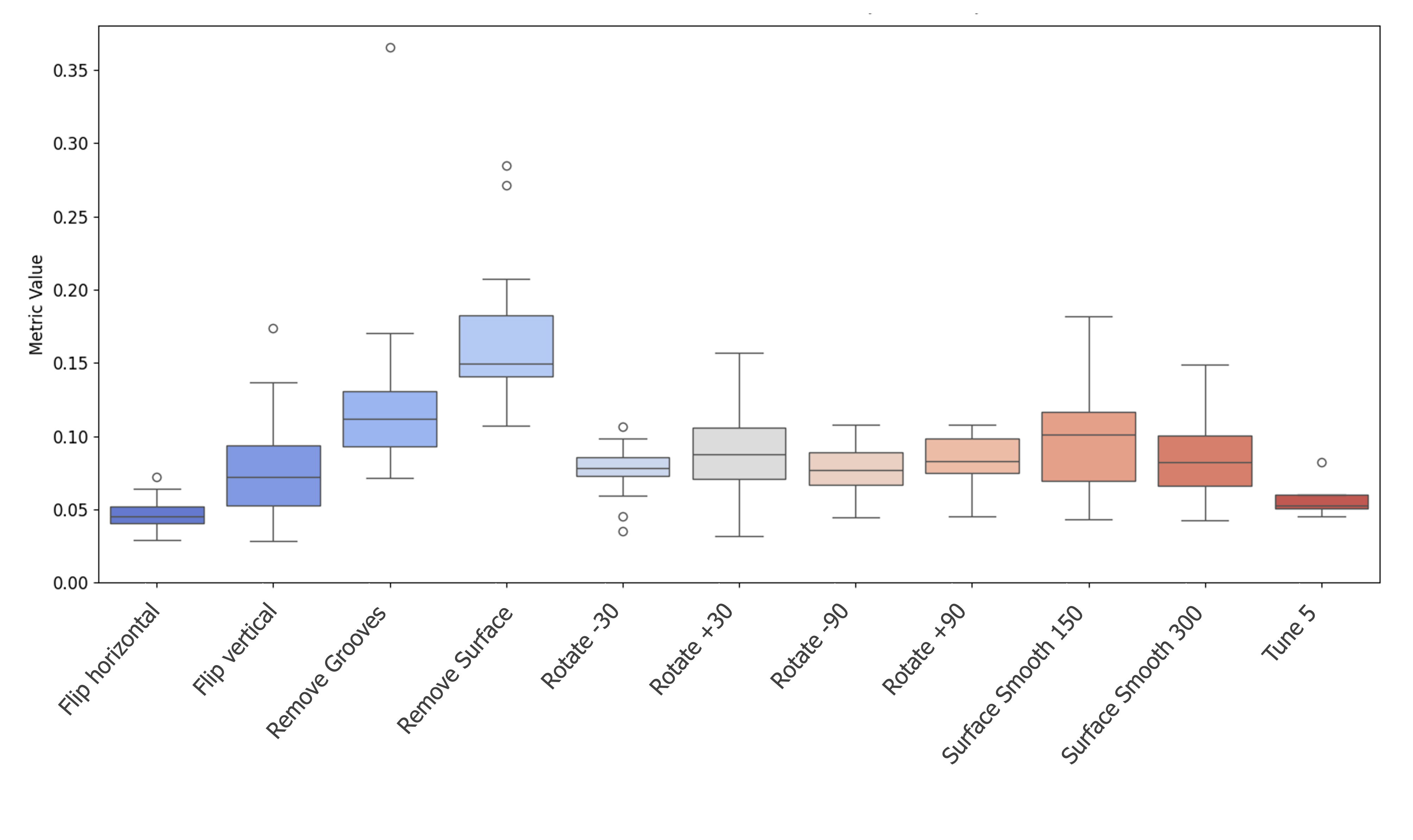} 
   \caption{Feature Importance of Operators on All Testing Images}
  \label{chart_barkXAI}
  \end{center}
\end{figure}

\subsection{Concept Mapping}
\label{concept_mapping}
For easier comparison with other XAI methods, we map our calculated concepts derived from operators into inferred concepts (Smooth, Plated, Rugged, Furrow, Vertical Stripped) which are more closely related to bark features by combining several calculated operations. We denote feature importance of a calculated concept $c$ with key $k$ and value $v$ as $FI(k, v)$ and significance of inferred concept as $Sig(c)$. We used values related to each operator and sort the inferred concepts from the most significant to the least significant based on significance values. Specifically, $Sig(smooth)$ is the maximum value between $FI({smooth}, +150)$ and $FI({smooth}, +300)$. $Sig(vertical\ stripped)$ is calculated as the average value of $FI({rotate}, -30)$ and $FI({rotate}, +30)$. $Sig(rugged)$ is the direct value of $FI(groove, remove)$. $Sig(plated)$ is the average value of $FI({rotate}, -30)$, $FI({rotate}, +30)$, $FI({flip}, horizontal)$ and $FI({flip}, vertical)$. $Sig(furrow)$ is the average value of $Sig(rugged)$ and $Sig(vertical\ stripped)$. The inferred concepts are then ordered based on the significance values. Please note, the way how the concept significance values is calculated is based on how each concept can be further decomposed and represented with the calculated concepts from operators, alternative interpretation can exist since perception of concept can always be subjective, due to the nature of concept-based XAI methods.

\subsection{Explanation Example}
\label{explanation_example}
\begin{figure}
    \centering
    \includegraphics[width=\linewidth]{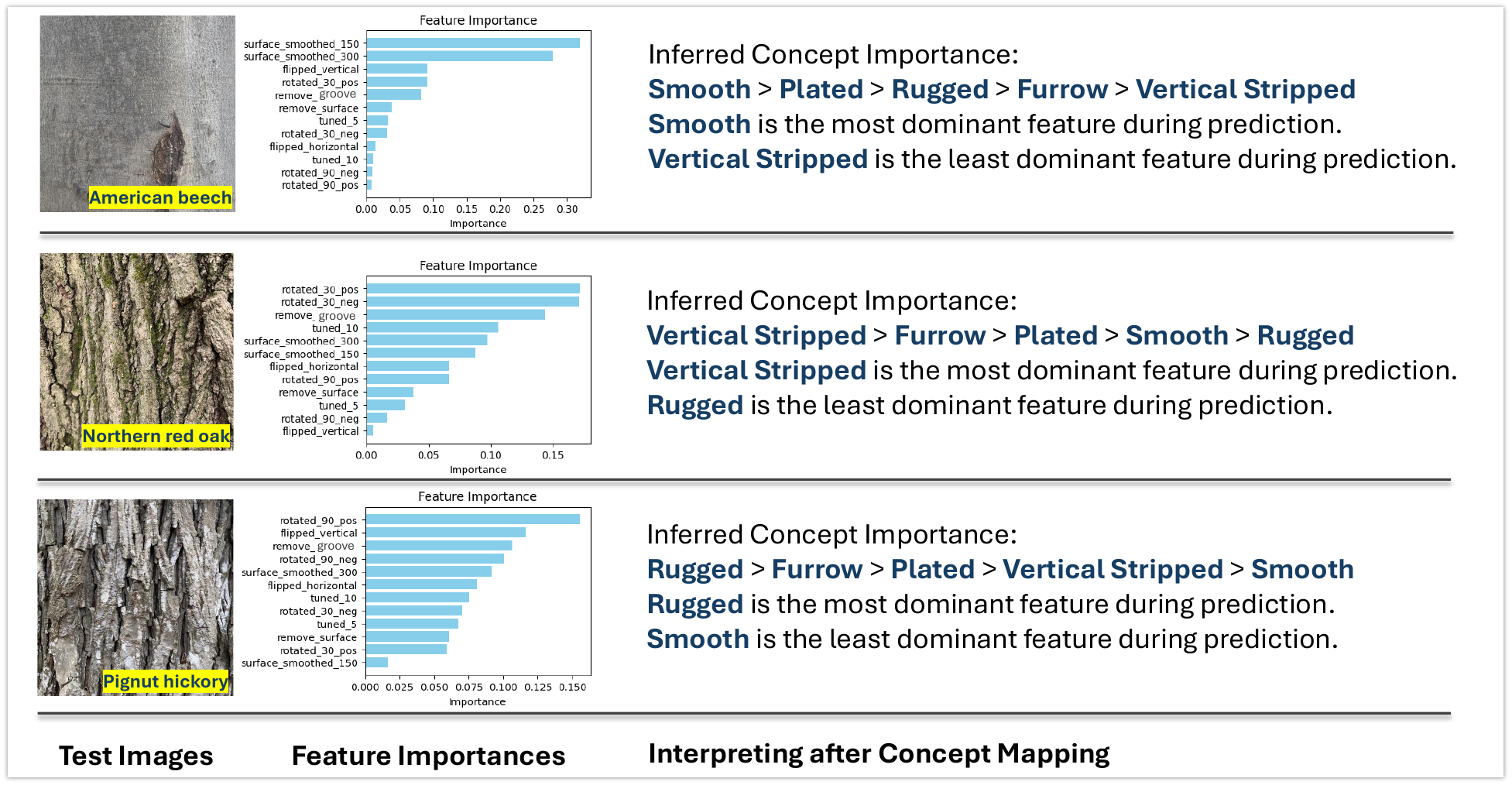}
    \caption{Examples of Explanations Generated using BarkXAI(LR). Feature importance values are calculated with softmax performed on slope values. The relative importance of inferred concepts are calculated based on feature importance.}
    \label{fig:example_explain}
\end{figure}

In Figure \ref{fig:example_explain}, we illustrate several explanation examples generated with BarkXAI(LR) on the trained vision model. Feature importance is calculated on each bark image and feature importance related to each calculated concept is shown. In the first image (American beech), the two most dominant calculated concept are both related to surface smooth, which indicates of the selected calculated concepts, smooth surface is the most important visual feature for the vision model to classify it as an American beech. In the second image (Northern red oak), the two most dominant calculated concept are both related to rotation. It shows the directionality (vertical) in the image is more important than others, such as color information (tuned\_10 and tuned\_5). The inferred concepts (Smooth, Plated, Rugged, Furrow, Vertical Stripped) are derived based on how they can be decomposed into the calculated concepts, which are used to facililate the comparison with other concept-based methods.

\begin{table}
\centering
\caption{Importance Rankings of Several Methods with Kendall’s Tau}
\label{Tab:tau_comparision}
\scalebox{0.8}{
\begin{tblr}{
  hline{1-2,23} = {-}{},
}
Species              & TCAV               & Llama3.2-Vision    & \textbf{BXAI (DT)}           & \textbf{BXAI (LR)}         & \textbf{BXAI (RF)}         \\
American basswood   & $-0.011 \pm 0.328$ & $-0.2 \pm 0.415 $  & $0.126 \pm 0.431 $  & $0.442 \pm 0.479$ & $0.063 \pm 0.454$ \\
American beech      & $-0.063 \pm 0.225$ & $0.032 \pm 0.441 $ & $0.484 \pm 0.375 $  & $0.863 \pm 0.146$ & $0.495 \pm 0.442$ \\
American elm        & $0.09 \pm 0.337 $  & $-0.12 \pm 0.515 $ & $0.47 \pm 0.365 $   & $0.83 \pm 0.255 $ & $0.44 \pm 0.459 $ \\
American sycamore   & $0.418 \pm 0.324 $ & $-0.236 \pm 0.389$ & $0.509 \pm 0.446 $  & $0.927 \pm 0.23 $ & $0.473 \pm 0.299$ \\
Bitternut hickory   & $-0.015 \pm 0.199$ & $0.108 \pm 0.32 $  & $0.354 \pm 0.438 $  & $0.631 \pm 0.57 $ & $0.477 \pm 0.404$ \\
Black cherry        & $-0.244 \pm 0.43 $ & $0.044 \pm 0.445 $ & $0.289 \pm 0.5 $    & $0.733 \pm 0.333$ & $0.456 \pm 0.498$ \\
Black oak           & $0.089 \pm 0.486 $ & $-0.1 \pm 0.423 $  & $0.367 \pm 0.477 $  & $0.722 \pm 0.406$ & $0.489 \pm 0.378$ \\
Black walnut        & $-0.14 \pm 0.254 $ & $0.08 \pm 0.325 $  & $0.16 \pm 0.28 $    & $0.66 \pm 0.269 $ & $0.18 \pm 0.34 $  \\
Eastern cottonwood  & $-0.117 \pm 0.264$ & $0.033 \pm 0.415 $ & $-0.017 \pm 0.436 $ & $0.567 \pm 0.446$ & $0.233 \pm 0.399$ \\
Eastern white pine & $-0.28 \pm 0.271 $ & $0.2 \pm 0.438 $   & $0.24 \pm 0.427 $   & $0.24 \pm 0.496 $ & $0.4 \pm 0.335 $  \\
Hackberry            & $-0.19 \pm 0.313 $ & $-0.12 \pm 0.354 $ & $0.26 \pm 0.415 $   & $0.55 \pm 0.46 $  & $0.31 \pm 0.462 $ \\
Honeylocust          & $0.022 \pm 0.457 $ & $0.022 \pm 0.346 $ & $-0.111 \pm 0.285 $ & $0.467 \pm 0.481$ & $0.022 \pm 0.416$ \\
Northern red oak   & $0.221 \pm 0.361 $ & $-0.2 \pm 0.486 $  & $0.189 \pm 0.483 $  & $0.347 \pm 0.415$ & $0.179 \pm 0.366$ \\
Pignut hickory      & $-0.07 \pm 0.376 $ & $-0.24 \pm 0.403 $ & $0.05 \pm 0.275 $   & $0.64 \pm 0.463 $ & $0.23 \pm 0.541 $ \\
Sassafras            & $0.046 \pm 0.491 $ & $-0.2 \pm 0.392 $  & $0.092 \pm 0.512 $  & $0.4 \pm 0.376 $  & $0.292 \pm 0.397$ \\
Shagbark hickory    & $-0.057 \pm 0.386$ & $0.171 \pm 0.345 $ & $0.133 \pm 0.438 $  & $0.667 \pm 0.442$ & $0.114 \pm 0.522$ \\
Silver maple        & $0.2 \pm 0.245 $   & $-0.35 \pm 0.218 $ & $0.35 \pm 0.384 $   & $0.6 \pm 0.316 $  & $0.45 \pm 0.456 $ \\
Sugar maple         & $0.19 \pm 0.306 $  & $0.03 \pm 0.324 $  & $-0.13 \pm 0.359 $  & $0.01 \pm 0.436 $ & $-0.04 \pm 0.463$ \\
White ash           & $0.088 \pm 0.36 $  & $0.05 \pm 0.357 $  & $0.225 \pm 0.429 $  & $0.437 \pm 0.459$ & $0.175 \pm 0.463$ \\
White oak           & $0.25 \pm 0.296 $  & $0.12 \pm 0.349 $  & $0.0 \pm 0.335 $    & $0.54 \pm 0.415 $ & $0.23 \pm 0.381 $ \\
Yellow poplar       & $0.12 \pm 0.299 $  & $-0.04 \pm 0.383 $ & $0.37 \pm 0.359 $   & $0.61 \pm 0.412 $ & $0.25 \pm 0.384 $ 
\end{tblr}
}
\end{table}

 \begin{figure}[h]
     \centering
    \includegraphics[width=0.9\textwidth, height=180pt]{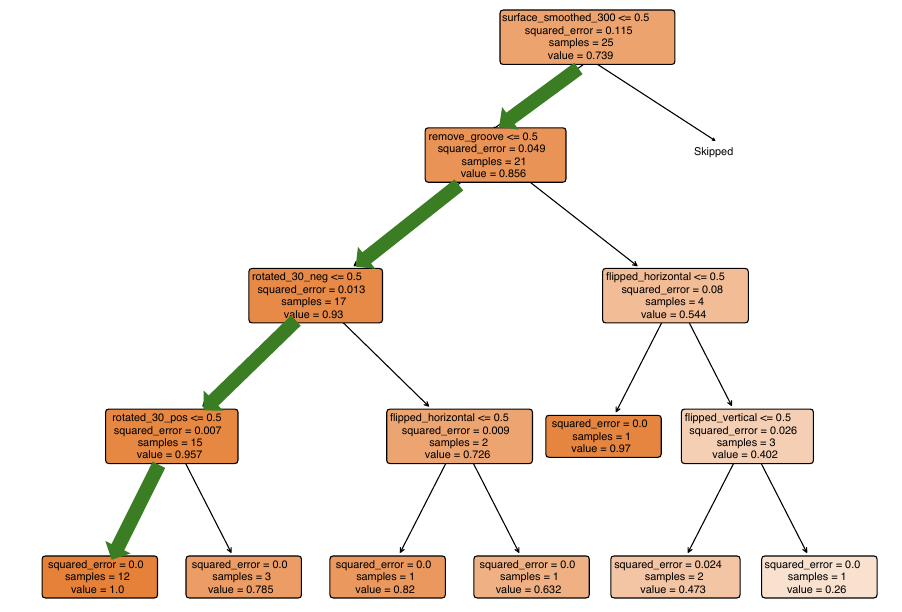} 
   \caption{Example of Using Decision Tree as Surrogate Model to Show Inter-Concept Relationships between Concepts. The path denoted by green arrows is the most likely path of the reasoning process of the vision model given the particular image input. This serves as an example of interpreting the vision model based on calculated concepts. The inter-concept relationship can be explored with decision tree.}
  \label{chart_barkXAI_TREE}
\end{figure}

\begin{table}
\centering
\caption{ Tree species with scientific name in the bark dataset.}
\label{21species_list}
\begin{tblr}{
  hline{1-2,13} = {-}{},
}
\textbf{Common Name} & \textbf{Species}               & \textbf{Common Name} & \textbf{Species}                \\
American basswood    & \textit{Tilia americana}       & Honey locust         & \textit{Gleditsia triacanthos}  \\
American beech       & \textit{Fagus grandifolia}     & Northern red oak     & \textit{Quercus~rubra}          \\
American elm         & \textit{Ulmus americana}       & Pignut hickory       & \textit{Carya glabra}           \\
American sycamore    & \textit{Platanus occidentalis} & Sassafras            & \textit{Sassafras albidum}      \\
Bitternut hickory    & Carya \textit{cordiformis}     & Shagbark hickory     & \textit{Carya ovata}            \\
Black cherry         & \textit{Prunus serotina}       & Silver maple         & \textit{Acer saccharinum}       \\
Black oak            & \textit{Quercus velutina}      & Sugar maple          & \textit{Acer saccharum}         \\
Black walnut         & \textit{Juglans nigra}         & White ash            & \textit{Fraxinus americana}     \\
~Eastern cottonwood  & \textit{~ ~Populus deltoides}  & White oak            & \textit{Quercus alba}           \\
Eastern white pine   & \textit{Pinus strobus}         & Yellow poplar        & \textit{Liriodendron tulipifer} \\
Hackberry            & \textit{Celtis occidentalis}   &                      &                                 
\end{tblr}
\end{table}

\clearpage
\begin{figure}[h]
\begin{center}
\includegraphics[width=0.98\textwidth]{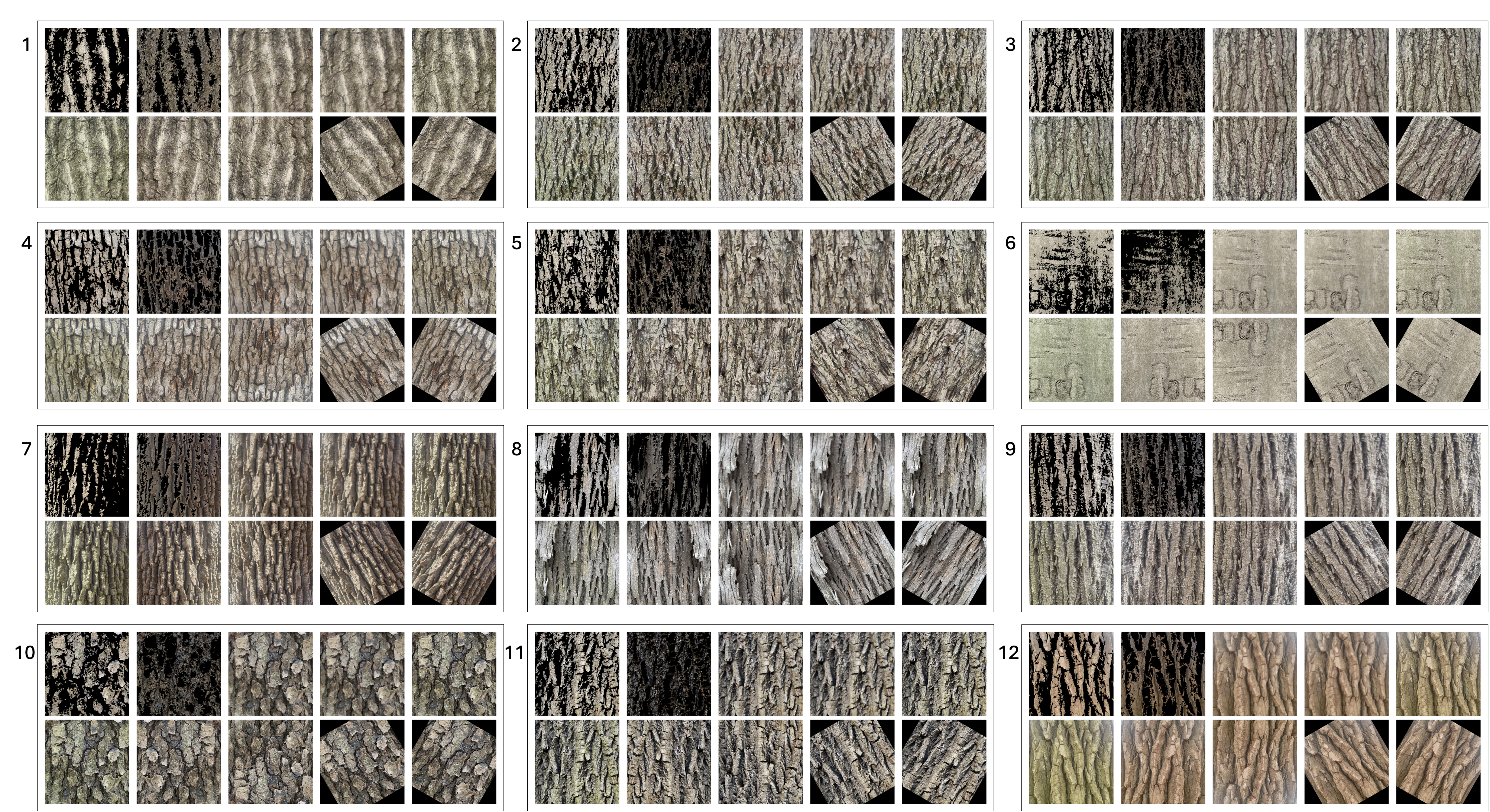} 
  \caption{Ten Operations Demonstration Examples Among 12 Species: (1. Northern red oak, 2. Bitternut hickory, 3. Sugar maple, 4. White oak, 5. Pignut hickory, 6. American beech 7. White ash, 8. Shagbark hickory, 9. American basswood, 10. Black cherry, 11. Black walnut, 12. Eastern cottonwood). }
\label{chart_barkXAI}
\end{center}
\end{figure}

\end{document}